\documentclass[journal]{IEEEtran}
\usepackage{amsmath,amsfonts}
\usepackage{algorithm}
\usepackage{algpseudocode}
\usepackage{array}
\usepackage[caption=false,font=normalsize,labelfont=sf,textfont=sf]{subfig}
\usepackage{textcomp}
\usepackage{url}
\usepackage{verbatim}
\usepackage{graphicx}
\usepackage{cite}
\usepackage{booktabs}

\algrenewcommand\algorithmicrequire{\textbf{Input:}}
\algrenewcommand\algorithmicensure{\textbf{Output:}}

\begin{document}

\title{DiverAge: Reliable Pluralistic Face Aging with Cross-Age Identity Relation Guidance}

\author{Yueying Zou, Peipei Li, Qianrui Teng, Dianyan Xu, Zekun Li 
\thanks{Yueying
Zou, Peipei Li, Qianrui Teng, and Dianyan Xu are with the School of Artificial Intelligence, Beijing University of Posts
and Telecommunications, Beijing 100876, China. E-mail: {zouyueying2001, lipeipei, qrteng, cocoxu}@bupt.edu.cn.}
\thanks{Zekun Li is with the School of Computer Science, University of California,
Santa Barbara, USA. E-mail: zekunli@cs.ucsb.edu.}
\thanks{Peipei Li is the corresponding author. E-mail: lipeipei@bupt.edu.cn.}}

\markboth{Journal of \LaTeX\ Class Files,~Vol.~14, No.~8, August~2021}%
{Shell \MakeLowercase{\textit{et al.}}: A Sample Article Using IEEEtran.cls for IEEE Journals}


\maketitle
\begin{figure*}
    \centering
    \includegraphics[width=\linewidth]{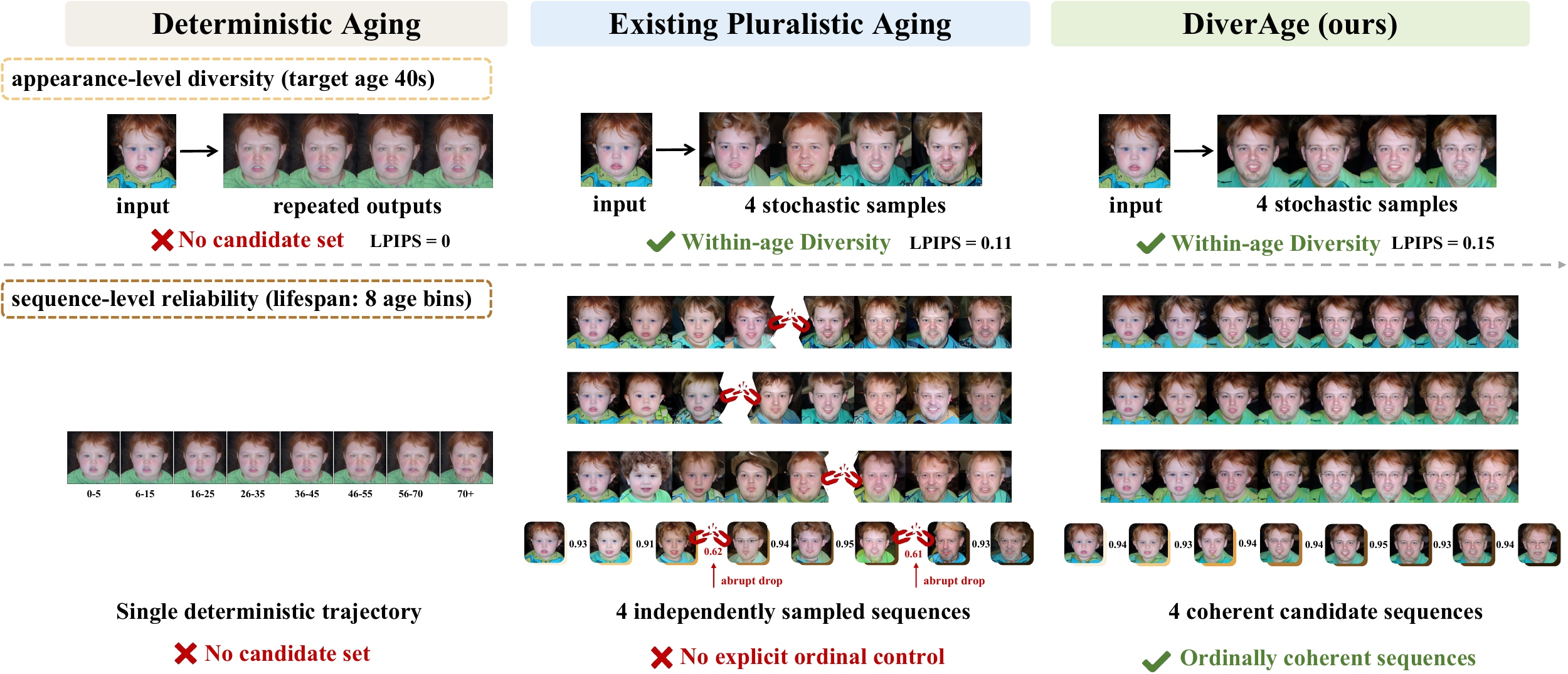}
    \caption{Overview of DiverAge motivation.
\textit{Left:} deterministic methods generate repeated outputs and fail to provide candidate diversity. 
\textit{Middle:} existing pluralistic methods produce stochastic samples at each target age, but independently sampled lifespan sequences may suffer from abrupt identity-similarity drops. 
\textit{Right:} DiverAge maintains within-age diversity while producing ordinally coherent candidate sequences across lifespan age bins.}
    \label{fig:placeholder}
\end{figure*}

\begin{abstract}
Face aging plays an important role in long-term biometric analysis, cross-age identity verification, and forensic identity analysis. Since the same subject may exhibit multiple plausible appearances at a target age due to genetic, environmental, and lifestyle factors, face aging is inherently a one-to-many generation problem. However, pluralism alone is insufficient for reliable face aging: a model should provide appearance-level candidate diversity within each age group while maintaining sequence-level ordinal reliability across ordered age groups. Existing deterministic aging methods can synthesize visually plausible age-progressed faces, but usually lack stochastic diversity. In contrast, pluralistic aging methods introduce local appearance variations, but often fail to explicitly regulate the identity evolution of the full aging sequence.
In this paper, we propose \textbf{DiverAge}, a hierarchical pluralistic face aging framework based on diffusion autoencoding. DiverAge preserves appearance-level diversity through stochastic diffusion decoding and age-conditioned semantic modulation. To improve sequence-level reliability, we introduce a Cross-age Identity Relation Regulator (CARR), an inference-time guidance strategy that jointly denoises multiple target age groups. CARR is guided by a Cross-age Identity Similarity (CIS) prior estimated from real same-identity cross-age pairs, and suppresses excessive cross-age identity drift through one-sided sampling-time guidance without modifying the training objective or introducing extra trainable parameters. Experiments demonstrate that DiverAge improves sequence-level ordinal reliability while maintaining identity preservation, age accuracy, image quality, and appearance-level diversity.
\end{abstract}

\section{INTRODUCTION}

Face aging is important for long-term biometric analysis~\cite{bestrowden2018longitudinal,li2018global, li2020hierarchical, li2020deep}, 
cross-age identity verification~\cite{huang2022mtlface, wang2023learning}, 
and forensic identity analysis~\cite{deb2019finding, teng2024exploring, li2019wavelet}. Unlike deterministic facial editing tasks, face aging is inherently pluralistic: the same subject may exhibit multiple plausible appearances at a target age due to genetic, environmental, and lifestyle factors~\cite{gunn2009why,krutmann2017skin,okada2013facial}. Therefore, reliable face aging should not be treated as a single-output problem, but as a one-to-many generation problem.

However, generating different-looking faces is not sufficient for reliable pluralistic aging. We argue that reliability should be considered at two levels: appearance-level candidate diversity and sequence-level ordinal reliability. The former requires plausible stochastic variations within the same input and target age, such as wrinkles, skin texture, and local pigmentation. The latter requires the generated age sequence to remain identity-consistent and to follow the natural ordinal structure of aging, where neighboring age groups are expected to be more similar than distant ones.

Existing face aging methods usually address only one side of this requirement. Diffusion-based age editing methods, such as FADING~\cite{chen2023face}, FaceTT~\cite{facett}, TimeMachine~\cite{timemachine}, Cradle2Cane~\cite{cradle2cane}, and AgeBooth~\cite{agebooth}, can generate visually plausible and identity-preserving age-progressed faces. However, under the same input and target age, they often produce highly similar outputs, and therefore lack the candidate diversity required by pluralistic aging. 
Pluralistic methods, including PADA~\cite{li2023pluralistic}, ADFD~\cite{adfd}, and Aging Multiverse~\cite{gong2025aging}, introduce stochasticity into the generation process and improve local appearance diversity. Yet their diversity is mainly reflected in wrinkles, texture, and other appearance-level details, while different target ages are usually generated without explicit sequence-level ordinal control. 
Another line of work, such as Lifespan~\cite{or2020lifespan}, DiffAge3D~\cite{diffage3d}, and MyTimeMachine~\cite{qi2024mytimemachine}, models aging along an ordered age axis. However, these methods typically map each identity to a single deterministic trajectory, leaving little room for pluralistic aging candidates. Therefore, existing methods either provide diverse appearances without reliable sequence ordering or ordered trajectories without sufficient pluralistic diversity.

This limitation is not fully captured by conventional evaluation protocols. Prior face aging studies evaluate image quality~\cite{heusel2017gans, zhang2024ssr}, aging accuracy~\cite{rothe2015dex}, and identity preservation~\cite{deng2019arcface,  narayan2025facexformer} on each generated image independently. These frame-level criteria are necessary, but they do not measure whether a generated lifespan sequence follows the ordinal identity-similarity pattern observed in real same-identity cross-age data. A method may produce plausible faces at individual target ages, yet still yield a sequence whose cross-age identity relationship is statistically unreliable. To make sequence-level ordinal reliability measurable, we estimate CIS prior from real same-identity cross-age pairs, which characterizes how identity similarity~\cite{deng2019arcface} naturally changes across ordered age groups.

Beyond evaluation, the CIS prior also provides a direct way to regularize the generation process. Based on this idea, we propose \textbf{DiverAge}, a hierarchical pluralistic face aging framework that addresses the above reliability problem within a diffusion autoencoding pipeline~\cite{preechakul2022diffusion}. DiverAge preserves appearance-level candidate diversity through stochastic diffusion decoding, while improving sequence-level ordinal reliability through an inference-time CARR. During DDIM~\cite{ddim} sampling, multiple target age groups are jointly denoised, and CARR uses the CIS prior to discourage excessive identity drift across age groups through one-sided sampling-time guidance~\cite{ho2022classifier, dhariwal2021diffusion}. Importantly, CARR does not modify the diffusion training objective and introduces no additional trainable parameters.

This paper is an extension of our previous method PADA~\cite{li2023pluralistic}. 
Apart from providing more in-depth analysis and more extensive experiments, this journal version extends PADA from within-age appearance diversity to sequence-level cross-age reliability. 
Specifically, it introduces reliability-aware lifespan trajectory modeling, real-data CIS/APR-based sequence evaluation, and CARR-based inference-time guidance for ordinally reliable aging generation.
Our contributions are summarized as follows:
\begin{itemize}
    \item We formulate reliable pluralistic face aging as a \emph{hierarchical reliability problem} that requires both appearance-level candidate diversity and sequence-level ordinal reliability, beyond conventional frame-level evaluation of image quality, aging accuracy, and identity preservation.

    \item We construct a CIS prior from real same-identity cross-age pairs, providing a statistical reference for how identity similarity naturally changes across ordered age groups, and introduce APR-related metrics for sequence-level evaluation.

    \item We propose DiverAge, equipped with an inference-time CARR, which injects the CIS prior into DDIM sampling to improve sequence-level reliability without adding a training-time loss or extra trainable parameters.

    \item Extensive experiments show that DiverAge improves sequence-level ordinal reliability while maintaining identity preservation, age accuracy, image quality, and appearance-level candidate diversity.
\end{itemize}
\section{RELATED WORK}

\subsection{Diffusion-Based Face Aging}

The dominant paradigm for face aging has gradually shifted from
StyleGAN-based latent editing~\cite{or2020lifespan,karras2020analyzing,sam}
to controllable generation with diffusion models~\cite{rombach2022ldm}.
Built upon this progress, recent face-aging methods have explored age-conditioned diffusion editing from different perspectives~\cite{wang2024stablegarment}. FADING~\cite{chen2023face} explicitly incorporates age as a generative condition and serves as an important diffusion-based aging baseline. FaceTT~\cite{facett} uses vision-language models to produce attribute-aware prompts for age editing, TimeMachine~\cite{timemachine} introduces decoupled cross-attention and latent age guidance, and Cradle2Cane~\cite{cradle2cane} improves full-lifespan transformation with efficient diffusion sampling. Another line of work focuses on personalized adaptation: SelfAge~\cite{selfage} fine-tunes LoRA adapters using multi-age reference images, while AgeBooth~\cite{agebooth} combines young- and old-specific personalized adapters for cross-age generation.
Although these methods substantially improve visual realism, age
controllability, and identity preservation, they mainly treat face aging
as a deterministic editing task. For a given identity and target age, most of them generate a single plausible output, leaving the inherent
one-to-many nature of facial aging insufficiently modeled.

\subsection{Pluralistic Face Aging}
\label{sec:plurality}

Pluralistic face aging assumes that face aging is not a deterministic mapping from an identity and a target age to a single output image, but a conditional distribution over plausible aged appearances. 
Early efforts pursued this view within GAN- and VAE-based frameworks. 
CAAE~\cite{zhang2017caae} pioneered conditional adversarial autoencoders for joint age progression and regression by traversing a learned age manifold; IPCGAN~\cite{wang2018ipcgan} introduced identity-preserved conditional GANs that enforce identity consistency via an auxiliary face recognition loss; 
PFA-GAN~\cite{huang2021pfagan} extended cGAN-based aging to a progressive multi-stage formulation that decomposes large age gaps into incremental sub-translations; HRFAE~\cite{yao2021hrfae} achieved continuous high-resolution age editing through an encoder-decoder with feature modulation; and RAGAN~\cite{makhmudkhujaev2021ragan} learned personalized re-aging by modeling high-order interactions between input identity and target age. 
Building on this line of work, AAD-GAN~\cite{aadgan} introduces a VAE-based age-specific encoder that samples diverse age codes to generate multimodal aging faces under a shared identity-irrelevant representation, a direction also explored in hierarchical disentanglement frameworks~\cite{li2020hierarchical}; ADFD~\cite{adfd} analyses the StyleGAN latent space and randomly perturbs channels that control age-dependent attributes such as wrinkles, hair volume, and face shape; and DLAT$^{+}$~\cite{dlatplus} decomposes lifespan age transformation into texture and shape sub-networks, with diversity mechanisms deployed in both branches under multiple consistency constraints. With the rise of diffusion models, PADA~\cite{li2023pluralistic} reformulates pluralistic aging within a CLIP-driven diffusion autoencoder: low-level stochastic details emerge from the iterative denoising process, while high-level diversity is captured by a Probabilistic Aging Embedding that represents age as a Gaussian distribution rather than a deterministic point. More recently, Aging Multiverse~\cite{gong2025aging} extends pluralistic aging to a training-free flow-matching framework on Flux-based DiT models and couples aging diversity with lifestyle-related factors, producing an ``aging tree'' of multiple plausible futures.

Despite these advances, existing pluralistic aging methods have so far focused on within-target-age appearance variation. None of them guarantees that the diverse outcomes generated for the same identity at successive target ages remain mutually coherent under a monotone aging progression. More fundamentally, pluralistic aging should not be reduced to generating more visual outcomes; it should yield reliable aging trajectories that exhibit both within-age appearance variation and cross-age ordinal consistency.

\subsection{Ordinally-Consistent Face Aging}
\label{sec:ordinal}

A complementary line of research, largely orthogonal to pluralistic aging, focuses on modeling continuous aging trajectories along the age axis for a given identity. Lifespan~\cite{or2020lifespan} maps an input face onto a continuous age axis anchored at six trained age groups; DiffAge3D~\cite{diffage3d} and subsequent 3D-aware frameworks~\cite{teng2024exploring} introduces 3D awareness to enable multi-view-consistent aging across the lifespan; MyTimeMachine~\cite{qi2024mytimemachine} leverages personal photo collections to fit identity-specific aging trajectories; and Cradle2Cane~\cite{cradle2cane} performs full-lifespan age transformation through a two-pass few-step diffusion framework and introduces a harmonic identity–age metric to jointly evaluate the identity–age trade-off at each target age. A common limitation of these methods is that \emph{each identity corresponds to a single aging trajectory}: either produced through a deterministic mapping from age inputs, or uniquely determined by a set of personal reference samples. As a consequence, they cannot represent the biological fact that the same individual may age along multiple plausible paths, and consequently cannot serve downstream tasks that require within-age phenotypic variation.

\subsection{Training-free Guidance for Diffusion Sampling}
A complementary line of work studies how to steer pretrained diffusion models at sampling time, without modifying the training objective. Classifier guidance~\cite{dhariwal2021diffusion} and classifier-free guidance~\cite{ho2022classifier} are the most common forms, but both require either a noise-aware classifier or condition-aware training. Recent training-free guidance methods relax this requirement: DPS~\cite{chung2023diffusion} uses a Tweedie estimate of 
$\hat{x}_0$ to compute likelihood gradients for general inverse problems; LGD~\cite{song2023loss} formulates plug-and-play loss-guided sampling with a Monte Carlo estimator; FreeDoM~\cite{yu2023freedom} constructs time-independent energy functions from off-the-shelf networks; Universal Guidance~\cite{bansal2023universal} generalizes classifier guidance to arbitrary differentiable losses without retraining; 
MPGD~\cite{he2024manifold} introduces a manifold-preserving shortcut and performs guidance in the latent space of pretrained autoencoders, and DOODL~\cite{wallace2023doodl} directly optimizes diffusion latents via an invertible diffusion process. Asyrp~\cite{kwon2023asyrp} further shows 
that the U-Net bottleneck of a pretrained diffusion model already exposes a usable semantic latent space (h-space) for editing.

\section{Sequence-level Ordinal Reliability}
\label{sec:ordinal_reliability}

Conventional face aging evaluation measures image quality, aging accuracy,
and identity preservation on each generated image independently~\cite{or2020lifespan, huang2022mtlface, gong2025aging, cradle2cane}. Although
these frame-level criteria are necessary, they do not capture whether a
generated lifespan sequence follows the identity-similarity pattern observed
in real same-identity cross-age data. To quantify this missing dimension, we
introduce a sequence-level reliability formulation based on real cross-age
identity statistics. We first construct a CIS prior from same-identity image pairs, and then define the APR score to measure how well a generated sequence follows this prior.
\subsection{Cross-age Identity Similarity Prior}
\label{subsec:cis_prior}

Let $\mathcal{D}=\{(I_{p,k}, a_{p,k})\}$ denote a cross-age face dataset~\cite{chen2014cacd, moschoglou2017agedb}, where $p$ indexes identity, $k$ indexes images of the same identity, and $a_{p,k}$ is the corresponding age. To estimate how identity similarity naturally changes with age, we enumerate all unordered same-identity image pairs for each subject:
\begin{equation}
\Omega_p=\{(I_{p,u}, I_{p,v}) \mid u < v\}.
\end{equation}
Only same-identity pairs are used, while cross-identity pairs are
excluded. This construction ensures that the resulting statistics
characterize the variation of the same identity under aging, rather
than inter-person differences.
Each image is assigned to one of $A=8$ ordered age groups:
$\mathcal{B}\!=\!\{[0,5],[6,15],[16,25],[26,35],[36,45],[46,55],[56,70],[71,100]\}$
The non-uniform bin widths reflect the non-linear visual rate of facial
aging and the sparsity of elderly samples in cross-age datasets. Let
$g(a)\in\{1,\ldots,A\}$ denote the age-bin index of age $a$.

We extract an $\ell_2$-normalized face-recognition embedding $f(I)$ for each image. In our implementation, we use an ArcFace-style recognition model~\cite{deng2019arcface} by default and also examine trends across multiple face-recognition systems (Face++\footnote{\url{https://www.faceplusplus.com}}, 
InsightFace~\cite{deng2019arcface}, 
and FaceNet~\cite{schroff2015facenet}) to assess stability. For a same-identity pair $(I_{p,u}, I_{p,v})$, its cross-age identity similarity is defined as
\begin{equation}
s_{uv}= f(I_{p,u})^\top f(I_{p,v}).
\end{equation}
For each pair of age groups $(i,j)$, we collect all same-identity pairs
whose ages fall into these two bins:
\begin{equation}
\Omega_{ij}
=
\{(I_{p,u}, I_{p,v}) \mid
g(a_{p,u})=i,\; g(a_{p,v})=j,\; u<v \}.
\end{equation}
The raw Cross-age Identity Similarity matrix is computed by averaging the recognition similarity within each age-bin pair:
\begin{equation}
S^{\mathrm{raw}}_{ij}
=
\frac{1}{|\Omega_{ij}|}
\sum_{(I_{p,u},I_{p,v})\in\Omega_{ij}}
f(I_{p,u})^\top f(I_{p,v}).
\end{equation}

The raw matrix is symmetrized and calibrated to reduce scale differences across datasets and recognition backbones. We then normalize the calibrated matrix by the maximum diagonal value:
\begin{equation}
Z_{\mathrm{CIS}}[i,j]
=
\frac{S^{\mathrm{comp}}_{ij}}{\alpha},
\qquad
\alpha = \max_i S^{\mathrm{comp}}_{ii}.
\end{equation}
The diagonal entries are set to one, i.e., $Z_{\mathrm{CIS}}[i,i]=1$. The resulting matrix $Z_{\mathrm{CIS}}\in\mathbb{R}^{A\times A}$ serves as a calibrated Cross-age Identity Similarity prior. Its off-diagonal entries encode the expected identity similarity of the same subject observed at two different age groups.

\begin{figure}[t]
    \centering
    \includegraphics[width=\linewidth]{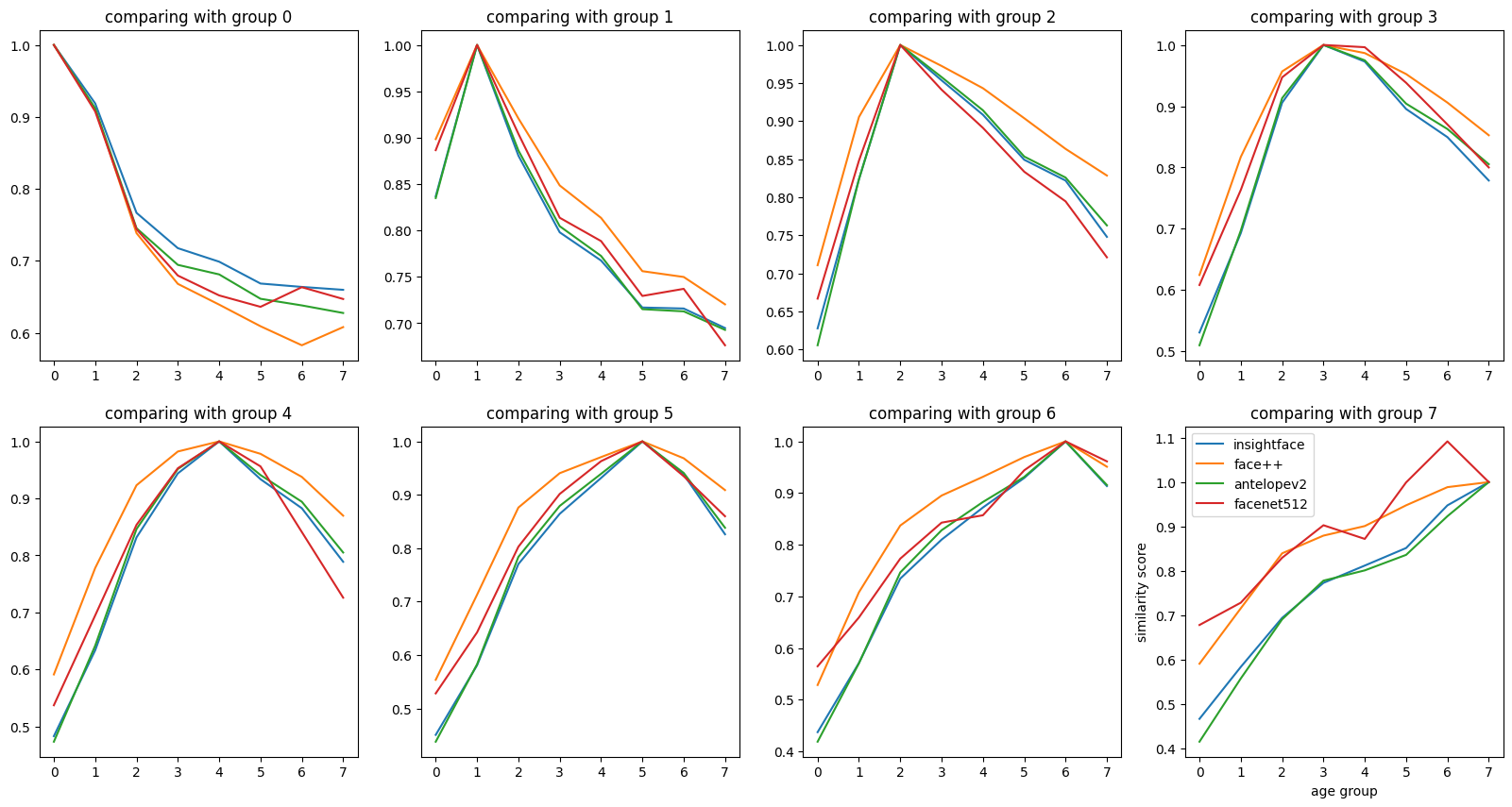}
    \caption{
    Real-data CIS trends estimated from same-identity cross-age pairs. Each panel fixes one reference age group and plots its average face-recognition similarity to all other age groups. The trends are consistent across different face-recognition systems, suggesting that same-identity similarity follows an ordinal age-dependent pattern. This statistical trend is used to construct the CIS prior $Z_{\mathrm{CIS}}$.
    }
    \label{fig:real_id_analysis}
\end{figure}

Fig.~\ref{fig:real_id_analysis} visualizes this real-data trend under
different face-recognition systems. Each sub-figure fixes one reference age group and plots its similarity to all other age groups. The curves show a consistent ordinal structure: identity similarity is highest around the reference group and generally decreases as the age gap increases, consistent with prior longitudinal and age-invariant face-recognition studies~\cite{park2010age, bestrowden2018longitudinal}.
We also observe that early age groups exhibit sharper
changes, which is consistent with the rapid facial development during childhood and adolescence~\cite{deb2019finding}. These observations motivate us to use $Z_{\mathrm{CIS}}$ as a population-level same-identity cross-age prior for evaluating and guiding face aging sequences.

\subsection{Age Progression Rationality Score}
\label{subsec:apr_score}

Given a generated age progression sequence $\hat{\mathcal{X}}=\{\hat{x}_1,\ldots,\hat{x}_A\}$, where $\hat{x}_i$ corresponds to the $i$-th target age group, we compute its generated cross-age identity similarity matrix:
\begin{equation}
\hat{S}_{ij}
=
f(\hat{x}_i)^\top f(\hat{x}_j),
\qquad
i,j\in\{1,\ldots,A\}.
\end{equation}
Here $f(\cdot)$ denotes the face-recognition embedding extractor used for evaluating identity similarity.

A direct way to evaluate sequence-level ordinal reliability is to
measure how far the generated similarity matrix $\hat{S}$ deviates from the real-data prior $Z_{\mathrm{CIS}}$. We define the CIS deviation as
\begin{equation}
\mathcal{E}_{\mathrm{CIS}}
=
\frac{1}{A(A-1)}
\sum_{i\ne j}
\left|
\hat{S}_{ij}
-
Z_{\mathrm{CIS}}[i,j]
\right|.
\end{equation}
A lower $\mathcal{E}_{\mathrm{CIS}}$ indicates that the generated lifespan sequence better follows the identity-similarity statistics of real same-identity aging data. For biometric applications, excessive identity drift is particularly harmful. We therefore also define a one-sided drift error:
\begin{equation}
\mathcal{E}_{\mathrm{drift}}
=
\frac{1}{A(A-1)}
\sum_{i\ne j}
\max\left(
0,\;
Z_{\mathrm{CIS}}[i,j] - \hat{S}_{ij}
\right).
\end{equation}
This term penalizes generated age pairs whose identity similarity falls
below the expected real-data prior, while leaving pairs with sufficient
similarity unchanged. This one-sided formulation directly reflects the
goal of preventing unrealistic identity drift during age progression.

We refer to the resulting sequence-level reliability measurement as the Age Progression Rationality (APR) score. Unlike conventional
identity-preservation metrics that compare each generated image only with the input image, APR evaluates the pairwise identity-similarity structure among all generated age groups. It therefore measures whether the full generated sequence follows the ordinal identity relationship observed in real same-identity cross-age data. In the following method section, the same CIS prior is further used as the sampling-time guidance prior for the proposed CARR.
\section{Method}
\label{sec:method}

\subsection{Overview}
\label{subsec:method_overview}

\begin{figure*}[t]
    \centering
    \includegraphics[width=\textwidth]{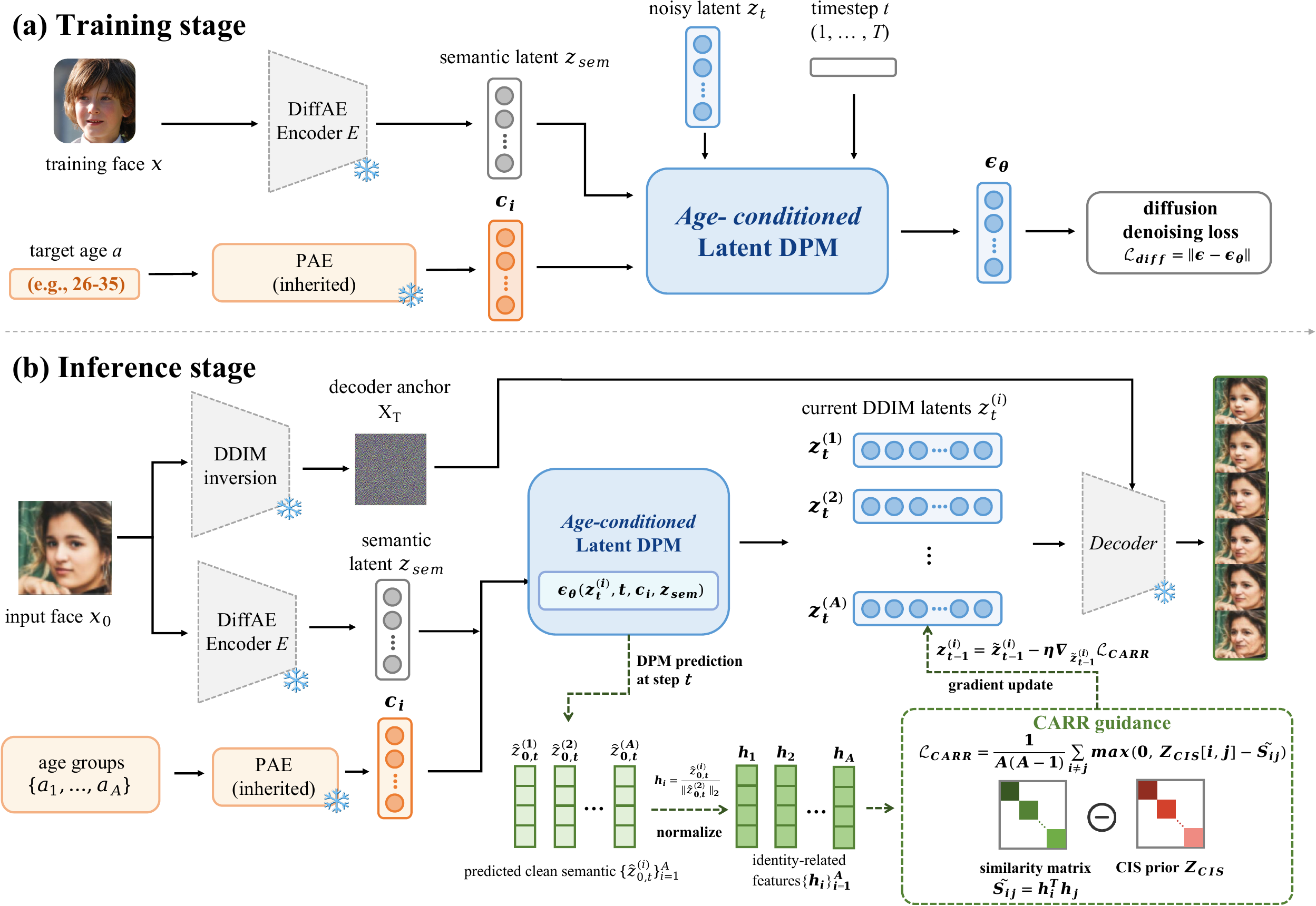}
    \caption{Architecture of DiverAge. (a)~Training stage: the age-conditioned latent diffusion prior is trained with the standard denoising objective. (b)~Inference stage: CARR jointly denoises all target age groups and applies CIS-guided gradient updates to improve sequence-level ordinal reliability.}
    \label{fig:method_overview}
\end{figure*}

DiverAge is designed around the hierarchical reliability formulation introduced in Sec .~\ref {sec:ordinal_reliability}. It aims to preserve appearance-level candidate diversity while improving sequence-level ordinal reliability. Given an input face image $x_0$ and an ordered set of target age groups $\mathcal{A}=\{a_1,\ldots,a_A\}$, DiverAge
generates an ordered age progression sequence
$\hat{\mathcal{X}}=\{\hat{x}_1,\ldots,\hat{x}_A\}$.

\begin{figure}[t]
    \centering
    \includegraphics[width=1\linewidth]{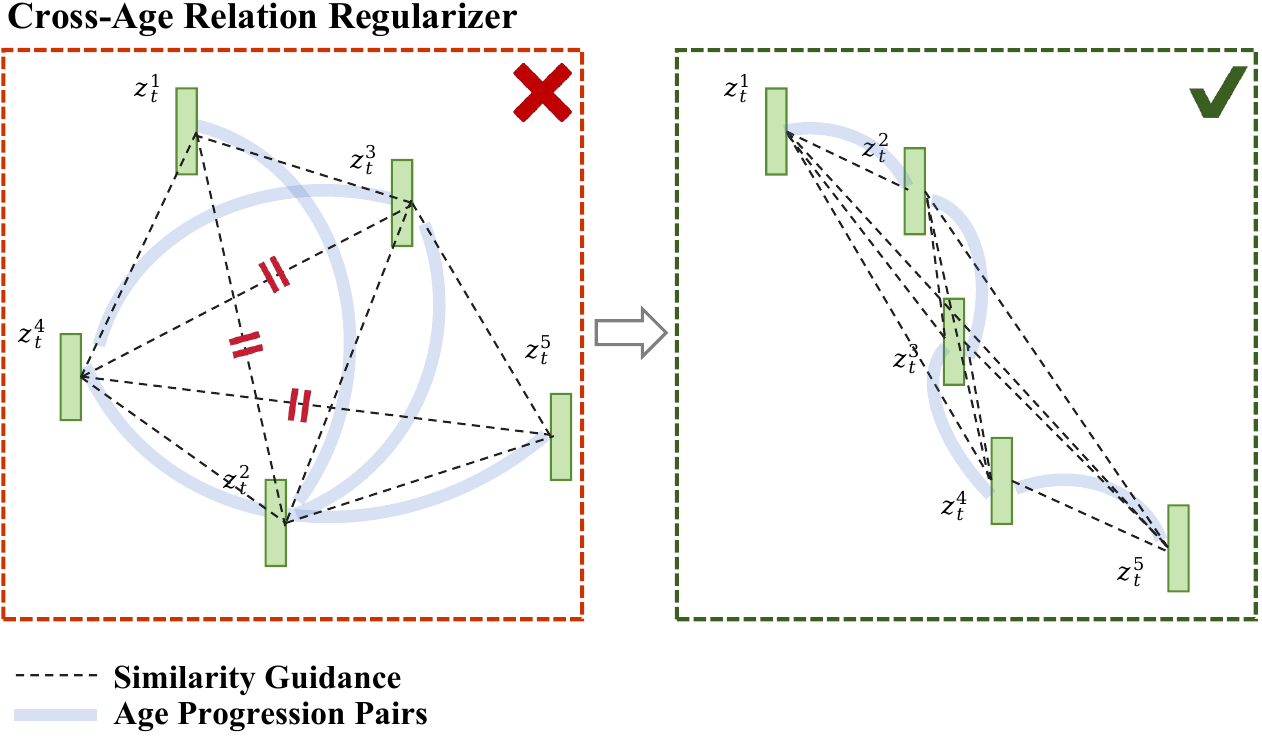}
    \caption{\textbf{Cross-Age Relation Regulator.}
    Nodes denote semantic codes of target age groups, and edges denote their pairwise cross-age similarities.
    CARR uses the CIS prior to guide DDIM sampling, reducing excessive identity drift while preserving valid cross-age relations.}
    \label{fig:cross_age_reg}
\end{figure}

The framework contains two levels. At the appearance level, DiverAge inherits stochastic candidate generation from the diffusion autoencoding backbone. This stochastic decoding process provides local appearance variations, such as wrinkles, skin texture, and subtle facial details, for the same input and target age. At the sequence level, DiverAge introduces an inference-time CARR(Fig.~\ref{fig:cross_age_reg}), which uses the CIS prior defined in Sec .~\ref {subsec:cis_prior} to discourage excessive identity drift across age groups during DDIM sampling. Fig.~\ref{fig:method_overview} illustrates the overall framework of DiverAge.

\subsection{Appearance-level Diversity via Diffusion Autoencoding}
\label{subsec:diffae_diversity}

DiverAge builds on a diffusion autoencoding backbone for
identity-preserving face editing. Given an input image $x_0$, a frozen encoder $E$ extracts a semantic identity latent:
\begin{equation}
z_{\mathrm{sem}} = E(x_0).
\end{equation}
Meanwhile, DDIM inversion maps the input image to a stochastic decoder anchor:
\begin{equation}
x_T = \mathrm{DDIMInv}(x_0).
\end{equation}
The semantic latent $z_{\mathrm{sem}}$ preserves identity-related information, while $x_T$ provides the stochastic decoding state used by the diffusion decoder.

Repeating the inference process with different stochastic decoder seeds or inversion noise produces multiple appearance-level candidates for the same input and target age. This diversity mainly appears in local aging details, such as skin texture, wrinkles, and pigmentation, rather than arbitrary high-level identity changes. This property is desirable for biometric-oriented aging: generated candidates should remain within a plausible neighborhood of the same identity while still reflecting appearance-level uncertainty.

\subsection{Age-conditioned Latent Diffusion Prior}
\label{subsec:latent_dpm}

To generate age-specific semantic latents, DiverAge employs an age-conditioned latent diffusion prior. For each target age group $a_i\in\mathcal{A}$, we first obtain an age condition vector:
\begin{equation}
c_i = \phi_{\mathrm{age}}(a_i),
\end{equation}
where $\phi_{\mathrm{age}}(\cdot)$ denotes the age embedding and MLP adapter.
Starting from a noisy latent $z_t^{(i)}$, the latent diffusion prior predicts the denoising direction conditioned on the timestep, the target age condition, and the semantic identity latent:
\begin{equation}
\epsilon_\theta^{(i)}
=
\epsilon_\theta
\left(
z_t^{(i)}, t, c_i, z_{\mathrm{sem}}
\right).
\end{equation}
A DDIM denoising step then updates the latent variable:
\begin{equation}
\tilde{z}_{t-1}^{(i)}
=
\mathrm{DDIMStep}
\left(
z_t^{(i)}, \epsilon_\theta^{(i)}, t
\right).
\end{equation}

For an age progression sequence, we denoise all target age groups jointly and synchronously. This produces a set of intermediate age-conditioned latents $\{\tilde{z}_{t-1}^{(i)}\}_{i=1}^{A}$ at each DDIM step, which will be regularized by CARR during inference.

\subsection{Cross-age Identity Relation Regulator}
\label{subsec:carr}

Although the age-conditioned latent diffusion prior can generate an ordered set of target-age results, different age groups may still be sampled without explicit control over their cross-age identity relationship. To improve sequence-level ordinal reliability, we propose CARR, an inference-time guidance mechanism driven by the CIS prior. Importantly, CARR is not used as a training-time loss: the latent diffusion prior is trained with the standard diffusion denoising objective, while CARR is applied only during inference as sampling-time guidance. Therefore, DiverAge can improve sequence-level ordinal reliability without introducing additional trainable parameters. An overview of the framework is shown in Fig.~\ref{fig:method_overview}.

At an intermediate DDIM step, CARR operates on the predicted semantic codes produced by the diffusion autoencoder. These codes are used as identity-related age-group representations rather than fully decoded images. We normalize the predicted semantic code of each age group and compute the generated cross-age similarity matrix:
\begin{equation}
\hat{S}_{ij}=h_i^\top h_j,
\end{equation}
where $h_i$ denotes the normalized semantic code of the $i$-th age group. The CIS prior used for CARR is calibrated to this semantic similarity space before guidance.

CARR then compares $\hat{S}$ with the calibrated CIS prior $Z_{\mathrm{CIS}}$. Since excessive identity drift is harmful for cross-age identity analysis, we adopt a one-sided hinge guidance:
\begin{equation}
\mathcal{L}_{\mathrm{CARR}}
=
\frac{1}{A(A-1)}
\sum_{i\ne j}
\max\left(0, Z_{\mathrm{CIS}}[i,j]-\hat{S}_{ij}\right).
\end{equation}
This loss penalizes age pairs whose generated identity similarity falls below the expected real-data prior, while leaving sufficiently similar pairs unchanged. The ordinal structure is encoded in $Z_{\mathrm{CIS}}$: age pairs with larger age gaps have lower expected similarity, while nearby age groups require stronger identity consistency.

CARR is applied as a sampling-time gradient guidance. Specifically, we update the current latent variable by
\begin{equation}
z_{t-1}^{(i)}
=
\tilde{z}_{t-1}^{(i)}
-
\eta
\nabla_{\tilde{z}_{t-1}^{(i)}}
\mathcal{L}_{\mathrm{CARR}},
\end{equation}
where $\eta$ is the guidance strength. This update is performed only during inference and does not modify the training objective of the latent diffusion prior.

\subsection{Inference Procedure}
\label{subsec:inference}
\begin{algorithm}[t]
\caption{Inference of DiverAge with CARR Guidance}
\label{alg:diverage_inference}
\begin{algorithmic}[1]
\Require Input face $x_0$, ordered target age groups
$\mathcal{A}=\{a_1,\ldots,a_A\}$, CIS prior $Z_{\mathrm{CIS}}$,
guidance strength $\eta$, DDIM steps $T$
\Ensure Age progression sequence
$\hat{\mathcal{X}}=\{\hat{x}_1,\ldots,\hat{x}_A\}$

\State $z_{\mathrm{sem}}\gets E(x_0)$
\State $x_T\gets \mathrm{DDIMInv}(x_0)$
\State $c_i\gets \phi_{\mathrm{age}}(a_i)$ for $i=1,\ldots,A$
\State Initialize $\{z_T^{(i)}\}_{i=1}^{A}$

\For{$t=T,\ldots,1$}
    \For{$i=1,\ldots,A$}
        \State $\epsilon_\theta^{(i)}
        \gets \epsilon_\theta(z_t^{(i)},t,c_i,z_{\mathrm{sem}})$
        \State $\tilde{z}_{t-1}^{(i)}
        \gets \mathrm{DDIMStep}(z_t^{(i)},\epsilon_\theta^{(i)},t)$
    \EndFor

    \If{CARR guidance is applied at step $t$}
        \State Compute normalized identity-related features
        $\{h_i\}_{i=1}^{A}$ from current predictions
        \State $\hat{S}_{ij}\gets h_i^\top h_j$
        \State $\mathcal{L}_{\mathrm{CARR}}
        \gets
        \frac{1}{A(A-1)}
        \sum_{i\ne j}
        \max(0, Z_{\mathrm{CIS}}[i,j]-\hat{S}_{ij})$
        \State $z_{t-1}^{(i)}
        \gets
        \tilde{z}_{t-1}^{(i)}
        -
        \eta\nabla_{\tilde{z}_{t-1}^{(i)}}
        \mathcal{L}_{\mathrm{CARR}}$
        for all $i$
    \Else
        \State $z_{t-1}^{(i)}\gets\tilde{z}_{t-1}^{(i)}$ for all $i$
    \EndIf
\EndFor

\State $\hat{x}_i\gets D(z_0^{(i)},x_T)$ for $i=1,\ldots,A$
\State \Return $\hat{\mathcal{X}}$
\end{algorithmic}
\end{algorithm}

During inference, DiverAge first extracts $z_{\mathrm{sem}}$ and obtains $x_T$ through DDIM inversion. For the ordered target age groups
$\mathcal{A}$, the corresponding age conditions $\{c_i\}$ are constructed. All target age groups are then jointly denoised. At selected sampling steps, CARR computes the generated similarity matrix and applies the CIS-guided update. After the final DDIM step, the frozen decoder maps the final latents to age-progressed images:
\begin{equation}
\hat{x}_i = D(z_0^{(i)}, x_T), \qquad i=1,\ldots,A.
\end{equation}
Repeating inference with different stochastic decoder seeds yields multiple appearance-level candidates for the same input.

\section{Experiments}
\label{sec:experiments}

We evaluate DiverAge according to the hierarchical reliability formulation proposed in this paper. Specifically, we aim to answer four questions: (1) whether DiverAge maintains standard frame-level aging
quality, including image quality, aging accuracy, and identity preservation; (2) whether it improves sequence-level ordinal reliability measured by APR/CIS-related errors; (3) whether it preserves appearance-level candidate diversity inherited from the pluralistic diffusion autoencoding backbone.

\begin{figure*}[t]
    \centering
    \includegraphics[width=1\linewidth]{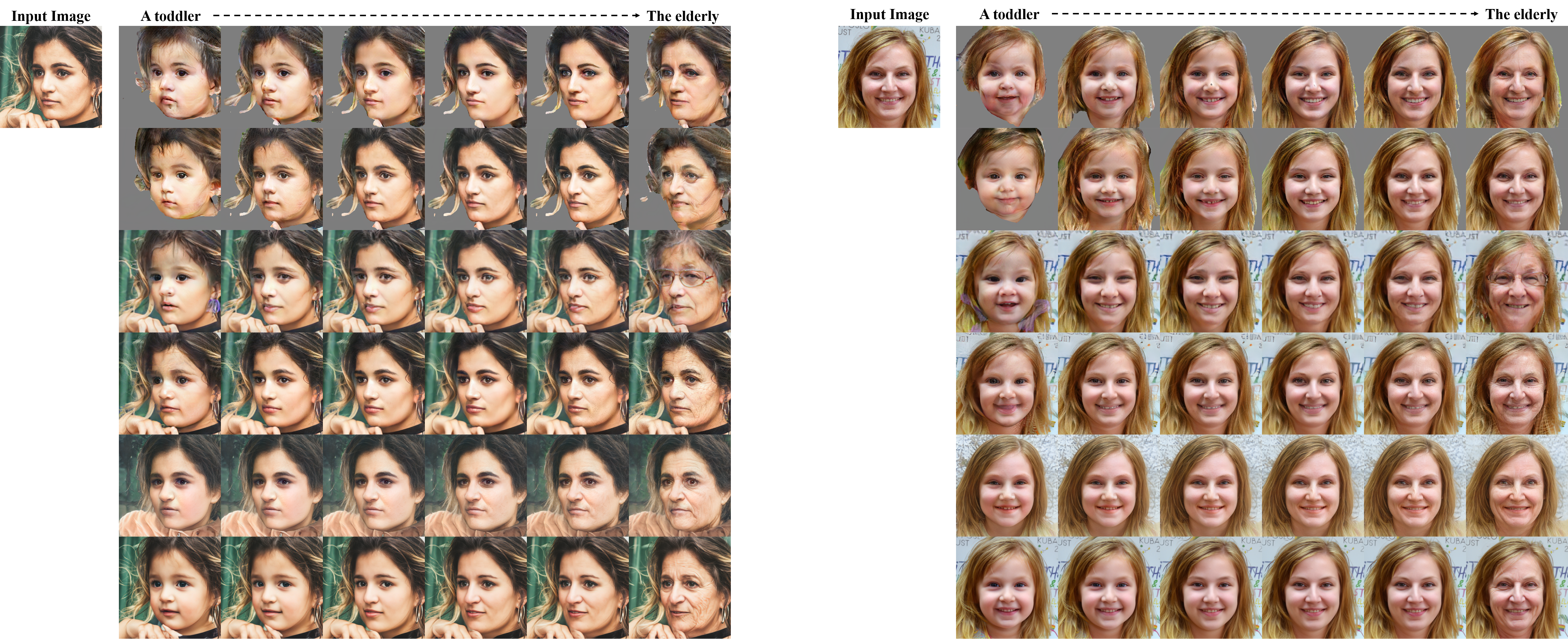}
    \caption{
   Qualitative comparison of age progression on FFHQ-AT.
Compared with deterministic baselines, DiverAge supports pluralistic generation and improves cross-age sequence reliability while preserving visual quality.
    }
    \label{fig:qualitative_comparison}
\end{figure*}

\begin{figure*}[h]
    \centering
    \includegraphics[width=1\linewidth]{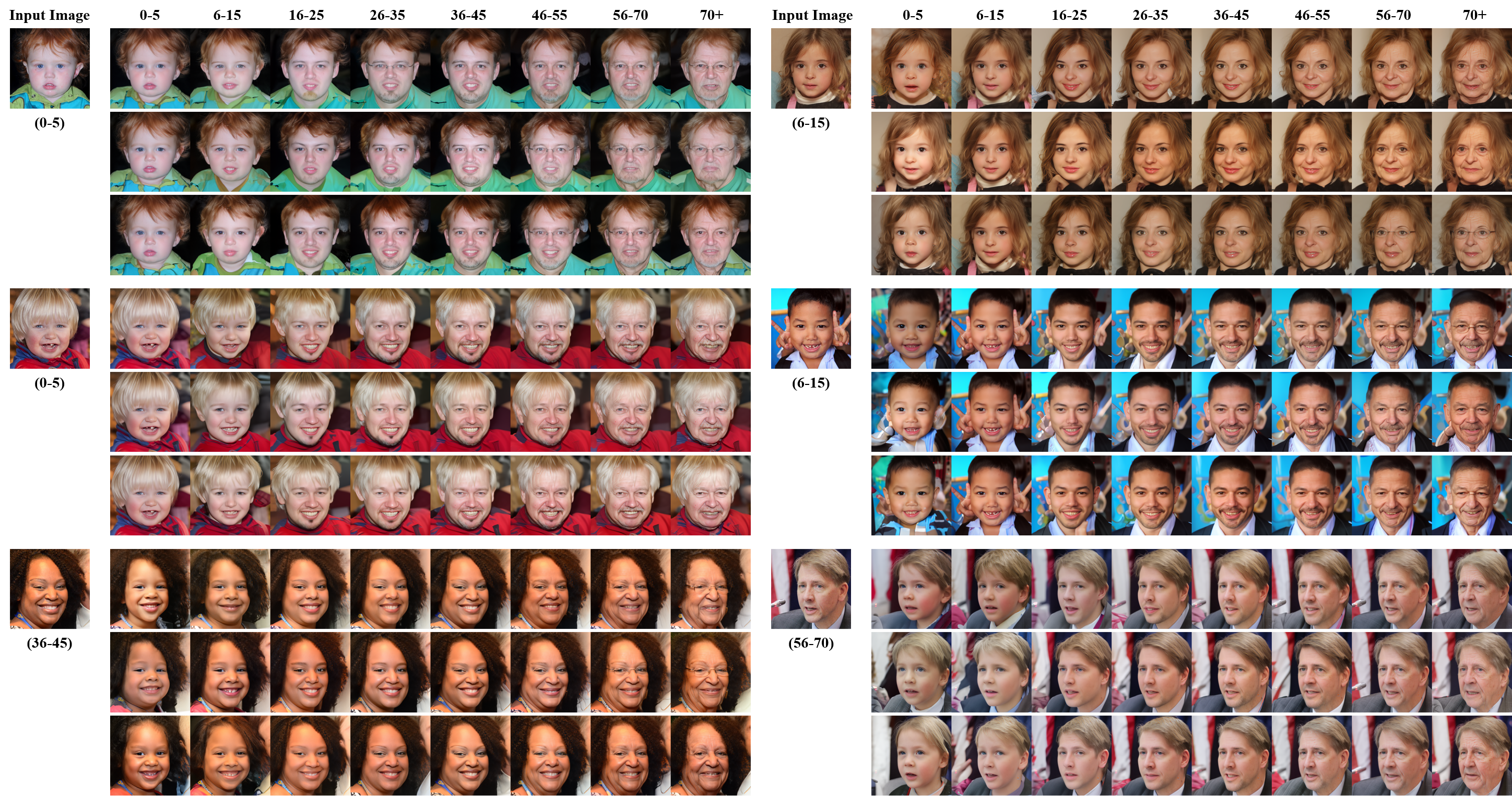}
    \caption{
    Pluralistic lifespan age progression results. Rows show different sampled aging sequences for the same input identity, and columns denote target age groups. DiverAge preserves appearance-level diversity while maintaining coherent cross-age progression.
    }
    \label{fig:lifespan_diversity}
\end{figure*}

\subsection{Experimental Setup}
\label{subsec:exp_setup}

\begin{figure*}[t]
    \centering
    \includegraphics[width=1\linewidth]{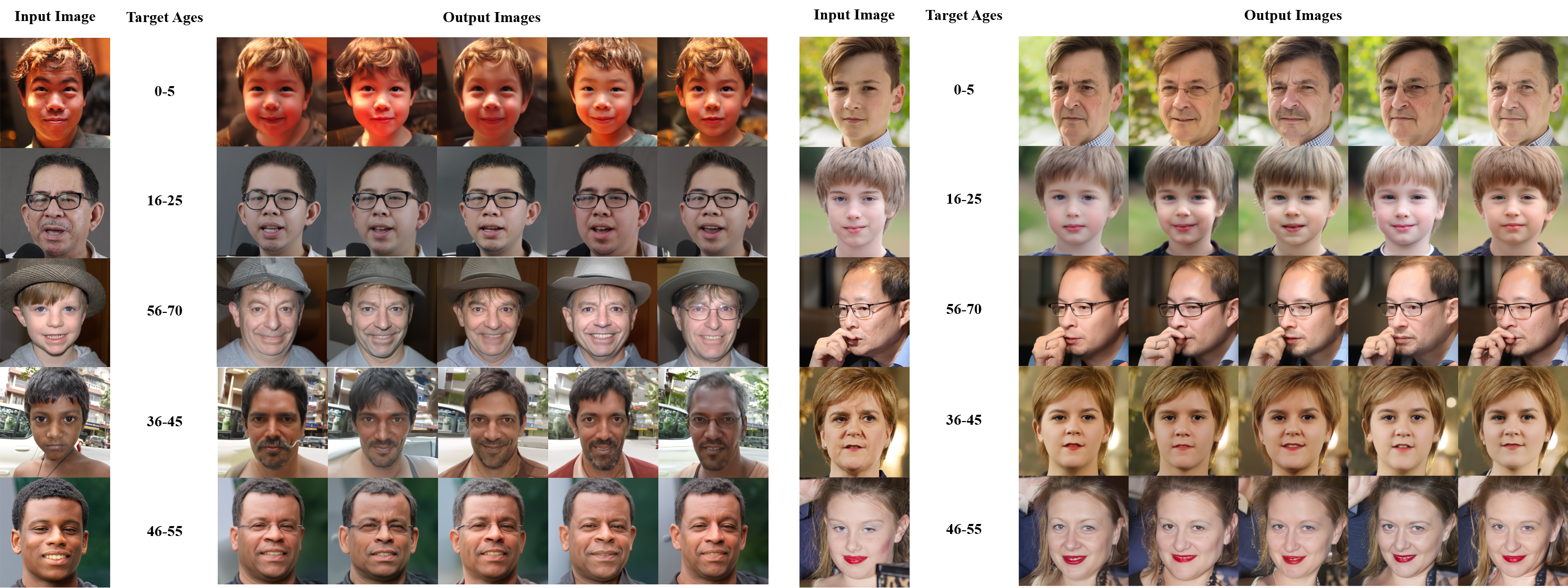}
    \caption{
    Appearance-level candidate diversity under the same input and target age. DiverAge generates multiple plausible candidates with local variations in wrinkles, skin texture, and other aging details, while preserving the target age and identity.
    }
    \label{fig:single_group_diversity}
\end{figure*}

\noindent\textbf{Datasets.}
Following PADA~\cite{li2023pluralistic}, we use FFHQ-AT as the main training and testing dataset for face aging evaluation. FFHQ-AT is constructed by relabeling FFHQ images into eight age groups: $0$--$5$, $6$--$15$, $16$--$25$, $26$--$35$, $36$--$45$, $46$--$55$, $56$--$70$, and $71+$, with gender-specific age-related text descriptions. A pretrained age estimator is used to assign each image to the closest predefined age group. We use 66,928 images for training and 3,072 images for testing, including 1,528 male and 1,544 female images. For sequence-level ordinal reliability, we use the CIS prior introduced in Sec.~\ref{sec:ordinal_reliability}. Unless otherwise specified, APR/CIS-based evaluation follows the same age-bin setting and identity feature extractor described in Sec.~\ref{subsec:cis_prior}.

\noindent\textbf{Baselines.}
We compare DiverAge with representative deterministic and pluralistic face aging methods, including LATS~\cite{or2020lifespan}, DLFS~\cite{dlfs}, SAM~\cite{sam}, CUSP~\cite{gomez2022custom}, AgeTransGAN~\cite{agetransgan}, and PADA~\cite{li2023pluralistic}. For fair comparison, we use official implementations and pretrained models when available. PADA is used as the inherited pluralistic diffusion-autoencoding baseline.

\noindent\textbf{Metrics.}
We evaluate the models from three perspectives. First, for frame-level aging quality, we report Age MAE for aging accuracy, face-recognition cosine similarity for identity preservation, and FID for image quality.
Second, for appearance-level candidate diversity, we compute LPIPS~\cite{zhang2018lpips} within groups of multiple samples generated from the same input and target age. Third, for sequence-level ordinal reliability, we report APR-related errors, including CIS deviation and identity drift error as defined in Sec.~\ref{subsec:apr_score}.

\noindent\textbf{Implementation Details.}
DiverAge builds upon the pretrained stochastic decoder and semantic encoder inherited from PADA. The PAE and age-conditioning modules are kept frozen. All images are resized to $256\times256$. The semantic diffusion model is trained with the standard denoising objective using Adam optimizer, a learning rate of $1\times10^{-4}$, and a batch size of $64$. During inference, we use $T_{\mathrm{eval}}=20$ DDIM steps with stochastic noise strength $0.2$. CARR is applied only in the last $4$ low-noise steps with guidance weight $10$, using the calibrated CIS prior.

\subsection{Main Comparison on Face Aging Quality}
\label{subsec:main_comparison}

We first compare DiverAge with existing face aging methods under the standard frame-level evaluation protocol. The goal of this experiment is not to evaluate pluralistic diversity, but to verify whether DiverAge maintains the basic single-image aging quality required by face aging, including age accuracy, identity preservation, and image quality. For a fair comparison with deterministic baselines, we use a single-output inference setting by setting the stochastic noise strength to zero and
fixing the sampling seed, so that one output is generated for each input-target-age pair.

\begin{table}[t]
\footnotesize
\centering
\caption{
Quantitative comparison of frame-level face aging performance on the FFHQ-AT test set. For a fair comparison with deterministic baselines, DiverAge is evaluated in a single-output inference setting, with stochastic noise strength set to zero and the sampling seed fixed.
}
\label{tab:frame_level_comparison}
\renewcommand{\arraystretch}{1}
\begin{tabular}{l|ccc}
\toprule
Method
& Age MAE $\downarrow$
& ID Sim. $\uparrow$
& FID $\downarrow$ \\
\midrule
LATS~\cite{or2020lifespan}
& 4.55 & 0.462 & 67.01 \\
DLFS~\cite{dlfs}
& 4.21 & 0.527 & 68.18 \\
SAM~\cite{sam}
& 4.78 & 0.421 & 41.31 \\
CUSP~\cite{gomez2022custom}
& \textbf{2.91} & 0.337 & 22.499 \\
AgeTransGAN~\cite{agetransgan}
& 6.73 & \textbf{0.583} & \textbf{17.99} \\
\midrule
DiverAge
& \underline{3.09} & \underline{0.543} & \underline{21.12} \\
\bottomrule
\end{tabular}
\end{table}

As shown in Table~\ref{tab:frame_level_comparison}, DiverAge achieves competitive frame-level performance in the non-diverse setting. Although CUSP obtains the lowest Age MAE and AgeTransGAN obtains the best FID and identity score, DiverAge achieves a balanced result across all three metrics. This result shows that introducing the proposed sequence-level reliability formulation does not compromise the basic requirements of face aging.

Fig.~\ref{fig:qualitative_comparison} shows qualitative comparisons across ordered age groups. DiverAge generates plausible age progression with reasonable shape deformation and texture transformation. These results verify DiverAge's single-output aging quality.

\subsection{Sequence-level Ordinal Reliability}
\label{subsec:seq_reliability}

The central claim of this paper is that pluralistic face aging
should not only generate diverse-looking candidates at a single
target age, but also produce ordinally reliable lifespan sequences. We therefore evaluate whether the generated cross-age identity-similarity structure follows the real CIS prior introduced in Sec.~\ref{subsec:cis_prior}.

Different from the single-target-age diversity evaluation in Sec.~\ref{subsec:appearance_diversity}, this experiment is conducted under a lifespan sequence setting. For each input identity, the model generates an ordered sequence covering all target age groups. APR Err. and IDAG are computed from the full generated sequence. When reporting LPIPS-div in this table, we compute sequence-level diversity by comparing different sampled lifespan sequences at corresponding age groups and then averaging the diversity over all target ages.

\begin{table}[t]
\scriptsize
\centering
\caption{Sequence-level ordinal reliability comparison.}
\label{tab:lifespan_reliability}
\setlength{\tabcolsep}{2.5pt}
\renewcommand{\arraystretch}{0.95}
\resizebox{\columnwidth}{!}{
\begin{tabular}{l|cccc}
\toprule
Method
& Age MAE $\downarrow$
& LPIPS-div $\uparrow$
& APR Err. $\downarrow$
& IDAG $\downarrow$ \\
\midrule
SAM~\cite{sam}
& 4.78 & -- & 0.1287 & 0.2721 \\
CUSP~\cite{gomez2022custom}
& \textbf{2.91} & -- & 0.0958 & 0.2759 \\
AgeTransGAN~\cite{agetransgan}
& 6.73 & -- & 0.1550 & 0.1707 \\
PADA~\cite{li2023pluralistic}
& 3.09 & -- & 0.0857 & 0.1567 \\
\midrule
\textbf{DiverAge}
& 3.16 & \textbf{0.0854} & \textbf{0.0562} & \textbf{0.1418} \\
\bottomrule
\end{tabular}
}
\end{table}

As shown in Table~\ref{tab:lifespan_reliability}, DiverAge achieves the lowest APR error among the compared methods, reducing sequence-level deviation from real same-identity cross-age statistics. Compared with PADA, DiverAge improves APR error from $0.0857$ to $0.0562$, while maintaining comparable Age MAE. This supports our key claim that CARR improves sequence-level ordinal reliability rather than merely increasing visual diversity.
Fig.~\ref{fig:lifespan_diversity} visualizes pluralistic lifespan age progression sequences. Within each row, the generated faces follow an ordered aging process across all target age groups. Across rows, different sampled sequences exhibit appearance-level variations. This qualitative behavior is consistent with our hierarchical reliability formulation: stochasticity is preserved at the candidate level, while
CARR encourages the full sequence to remain ordinally reliable.

\begin{figure}[h]
    \centering
    \includegraphics[width=0.7\linewidth]{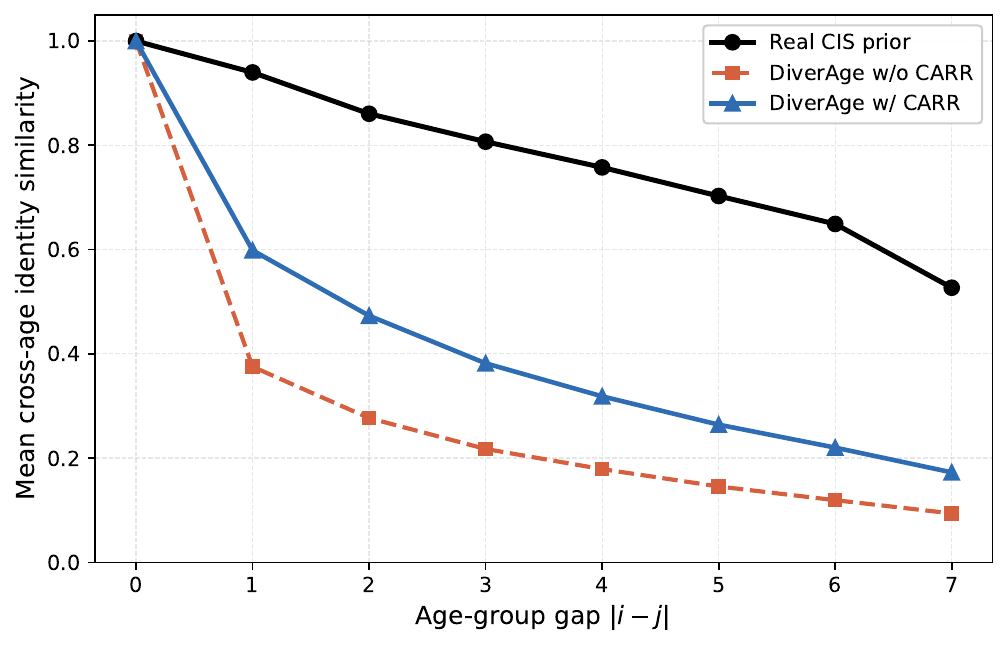}
    \vspace{-1pt}
    \caption{
    Cross-age identity similarity decay.
    }
    \label{fig:cis_decay}
\end{figure}

To further analyze whether CARR improves ordinal reliability rather than simply increasing identity similarity, we plot the mean cross-age identity similarity as a function of the age-group gap $|i-j|$ in Fig.~\ref{fig:cis_decay}. The real CIS prior shows a gradual decay pattern: neighboring age groups should remain highly identity-consistent, while larger age gaps naturally allow stronger identity variation.
Without CARR, the generated curve decays too sharply, indicating excessive identity drift even between nearby age groups. After applying CARR, the curve moves closer to the real CIS prior, especially for small and medium age gaps.
This demonstrates that CARR does not merely pull all age groups toward the same identity representation; instead, it calibrates the generated cross-age similarity structure toward the ordinal pattern observed in real same-identity aging data.

\subsection{Appearance-level Candidate Diversity}
\label{subsec:appearance_diversity}

We next evaluate whether DiverAge preserves appearance-level candidate diversity under a single-target-age setting. Unlike
Table~\ref{tab:lifespan_reliability}, which evaluates the full lifespan sequence across all age groups, fixes the input image and a target age, and generates multiple candidates with different stochastic seeds. We randomly select 500 test images, sample one target age group for each input, and generate 10 candidates per input.
As shown in Fig.~\ref{fig:single_group_diversity} and
Table~\ref{tab:appearance_diversity}, DiverAge produces multiple
plausible candidates for the same input and target age. The within-age LPIPS-div score increases compared with PADA-Diversity, while Age MAE and IDAG remain comparable. This indicates that DiverAge preserves appearance-level stochasticity without causing excessive identity drift or target-age mismatch.

\begin{table}[t]
\footnotesize
\centering
\caption{
Appearance-level candidate diversity under the same input and target age. 
}
\label{tab:appearance_diversity}
\renewcommand{\arraystretch}{1}
\begin{tabular}{l|ccc}
\toprule
Method
& LPIPS-div $\uparrow$
& Age MAE $\downarrow$
& IDAG $\downarrow$ \\
\midrule
PADA-Diversity~\cite{li2023pluralistic}
& 0.0297 & 3.388 & 0.1513 \\
DiverAge
& 0.1091 & 3.352 & 0.1509 \\
\bottomrule
\end{tabular}
\end{table}


\subsection{Ablation Study}
\label{subsec:ablation}

We conduct ablation studies to analyze the contribution of each component in DiverAge. This experiment is an internal progressive component analysis. All variants are evaluated under a unified lifespan sequence setting, where each input identity is used to generate an ordered sequence over all target age groups. This setting differs from the single-output frame-level evaluation in Table~\ref{tab:frame_level_comparison} and the single-target-age diversity evaluation in Table~\ref{tab:appearance_diversity}.
We follow the same implementation protocol as in Sec.~\ref{subsec:exp_setup}, using guidance strength $\eta=10$
and stochastic noise strength $0.2$ for CARR-based inference.

\begin{table*}[t]
  \footnotesize
  \centering
  \caption{
  Progressive component analysis of DiverAge. Each row adds one component on top of the previous one, and the last row corresponds to the full DiverAge model. 
  }
  \label{tab:ablation}
  \renewcommand{\arraystretch}{1}
  \begin{tabular}{l|ccc|ccc}
\toprule
Variant
& FID $\downarrow$
& Age MAE $\downarrow$
& ID Sim. $\uparrow$
& Seq. LPIPS-div $\uparrow$
& APR Err. $\downarrow$
& Drift Err. $\downarrow$ \\
\midrule
Inherited PADA backbone
& 25.23 & 6.04 & 0.252 & 0.459 & 0.300 & 0.393 \\
+ Age-conditioned latent prior
& 30.56 & 3.50 & 0.053 & 0.442 & 0.409 & 0.586 \\
+ Ordinal ID guidance
& 22.33 & 2.94 & 0.341 & 0.465 & 0.271 & 0.306 \\
+ CARR (full)
& \textbf{19.94} & \textbf{2.73} & \textbf{0.450} & \textbf{0.477} & \textbf{0.128} & \textbf{0.210} \\
\bottomrule
\end{tabular}
  \end{table*}

\begin{figure}[t]
    \centering
    \includegraphics[width=1\linewidth]{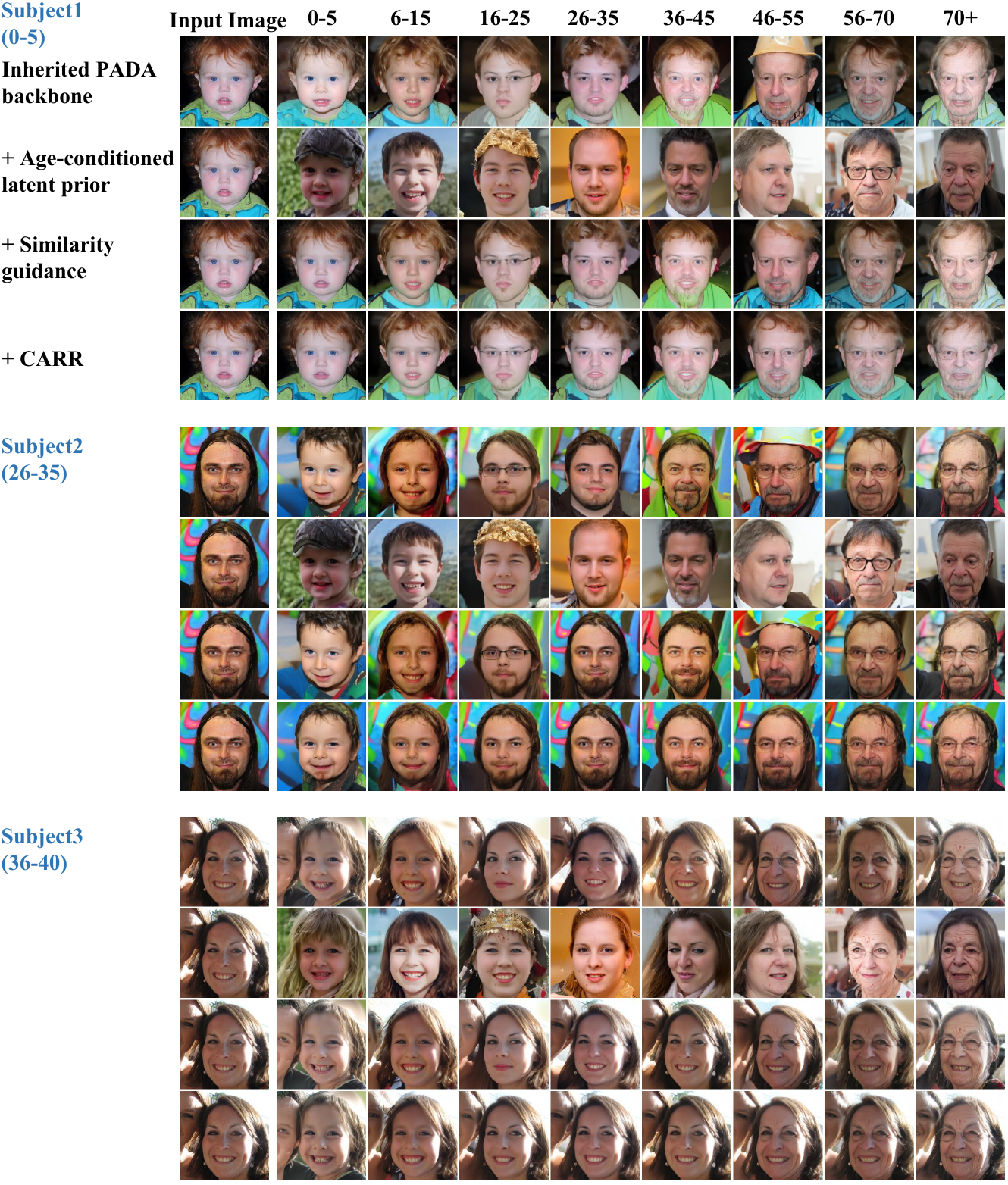}
    \caption{Qualitative results of progressive component analysis.}
    \label{fig:components_ablation}
\end{figure}

Quantitative results are summarized in Table~\ref{tab:ablation}, and qualitative comparisons are shown in Fig.~\ref{fig:components_ablation}.
This ablation is an internal progressive component analysis. The ``PADA'' row denotes the inherited diffusion-autoencoding backbone with synthesized age embeddings, without the age-conditioned latent prior or sampling-time guidance. It can
produce age-related changes, but lacks explicit control over
age-conditioned latent sampling and cross-age sequence reliability.

Adding the age-conditioned latent prior improves age controllability, as reflected by the reduced Age MAE and clearer age transformations in Fig.~\ref{fig:components_ablation}. However, without identity-relation guidance, the sampled semantic latents may drift from the source identity, leading to degraded ID Sim. and larger APR/Drift errors.
Naive ordinal ID guidance improves identity consistency by enforcing a monotonic similarity trend, but it does not calibrate cross-age relations to real lifespan statistics. With the CIS-calibrated CARR, the full model achieves the best overall trade-off across image quality, age controllability, identity preservation, diversity, and sequence-level reliability, producing more coherent cross-age identity evolution in Fig.~\ref{fig:components_ablation}.

\begin{figure}[t]
    \centering
    \includegraphics[width=0.7\linewidth]{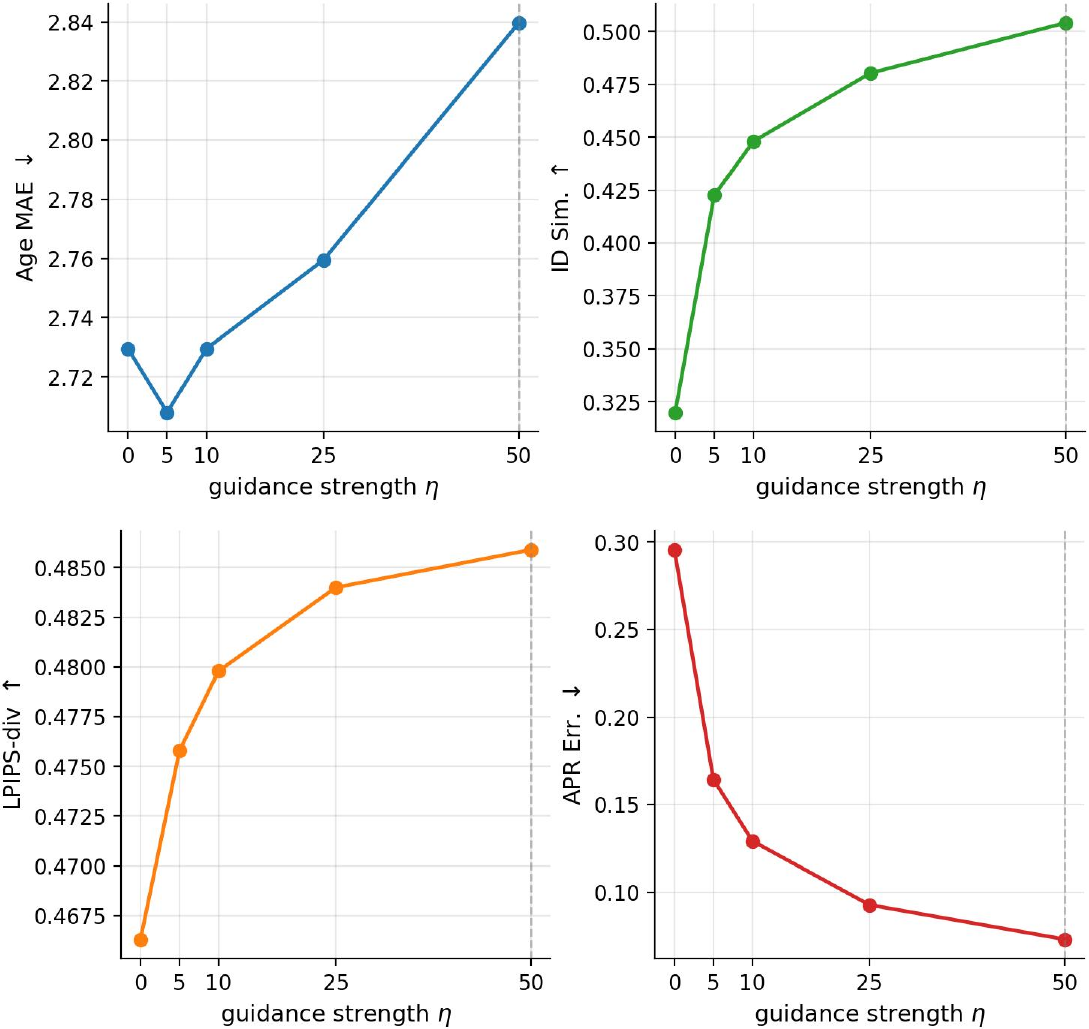}
    \caption{
    Sensitivity analysis of the CARR guidance strength $\eta$. }
    \label{fig:guidance_strength}
\end{figure}

\textbf{Guidance strength.}
We further study the effect of the CARR guidance strength $\eta$ on a $500$-source subset sampled from the FFHQ-256 test split (Fig.~\ref{fig:guidance_strength}). As $\eta$ increases, CARR consistently improves sequence-level reliability and identity preservation, showing that cross-age relation guidance is effective for correcting unstable aging trajectories. However, overly strong guidance starts to constrain the generated faces too much, leading to worse age controllability. We therefore set $\eta\!=\!10$ as the default, which provides a good balance between sequence reliability, identity preservation, and age accuracy.

\vspace{-2pt}

%
%


\section{Limitations and Discussion}
\label{sec:limitations}

Although DiverAge improves sequence-level ordinal reliability while preserving appearance-level candidate diversity, limitations
remain.
First, the CIS prior is estimated from face-recognition embeddings and may inherit biases or failure modes from the recognition backbone. Future work can explore demographic-aware or ensemble-based CIS priors for more robust cross-age reliability estimation.
Second, CARR is applied as inference-time guidance. While it introduces no trainable parameters and does not modify the diffusion training objective, it requires extra gradient computation during DDIM sampling. More efficient sequence-level guidance is worth exploring.
Third, DiverAge mainly focuses on facial aging. Contextual attributes such as hairstyle, clothing, background, and accessories may not evolve consistently with age. Extreme age groups, especially very young and very old faces, also remain challenging due to limited data. Extending DiverAge toward more controllable and context-aware aging trajectories is a future direction.
\section{Conclusion}
\label{sec:conclusion}

In this paper, we presented DiverAge, a hierarchical pluralistic face aging framework that jointly considers appearance-level candidate
diversity and sequence-level ordinal reliability. We constructed a Cross-age Identity Similarity (CIS) prior from real same-identity
cross-age pairs and introduced sequence-level metrics to evaluate whether generated aging trajectories follow realistic cross-age identity
relations.

DiverAge preserves stochastic appearance variation through diffusion autoencoding and further introduces CARR. This inference-time CIS-guided regulator aligns generated cross-age identity relations with the empirical prior during DDIM sampling. Experiments show that DiverAge
improves ordinal reliability while maintaining identity preservation, age accuracy, image quality, and appearance-level diversity. These results suggest that modeling pluralistic face aging as reliable trajectory generation is a promising direction for long-term cross-age identity analysis.

\clearpage

\bibliographystyle{IEEEtran}

\bibliography{IEEEabrv,egbib}

\end{document}